\documentclass[10pt,twocolumn,letterpaper]{article}

\usepackage{cvpr}
\usepackage{times}
\usepackage{epsfig}
\usepackage{graphicx}
\usepackage{amssymb}
\usepackage{helvet}  
\usepackage{courier}  
\usepackage{url}  
\usepackage[cmex10]{amsmath}
\usepackage{subfigure}
\usepackage[ruled,linesnumbered]{algorithm2e}
\usepackage{multirow}
\usepackage{diagbox}
\usepackage{flushend}
\usepackage[toc,page]{appendix}
\usepackage{xcolor}

\usepackage[pagebackref=true,breaklinks=true,letterpaper=true,colorlinks,bookmarks=false]{hyperref}
\graphicspath{ {images/pdf} }

\cvprfinalcopy
\begin{document}


%
\title{On Configurable Defense against Adversarial Example Attacks}

\author{Bo Luo, Min Li, Yu Li, Qiang Xu\\
The Chinese University of Hong Kong\\
\{boluo, mli, yuli, qxu\}@cse.cuhk.edu.hk
}
\maketitle
\begin{abstract}

Machine learning systems based on deep neural networks (DNNs) have gained mainstream adoption in many applications. Recently, however, DNNs are shown to be vulnerable to adversarial example attacks with slight perturbations on the inputs. Existing defense mechanisms against such attacks try to improve the overall robustness of the system, but they do not differentiate different targeted attacks even though the corresponding impacts may vary significantly. To tackle this problem, we propose a novel configurable defense mechanism in this work, wherein we are able to flexibly tune the robustness of the system against different targeted attacks to satisfy application requirements. This is achieved by refining the DNN loss function with an attack sensitive matrix to represent the impacts of different targeted attacks. Experimental results on CIFAR-10 and GTSRB data sets demonstrate the efficacy of the proposed solution.

\end{abstract}
\section{Introduction}

Deep neural networks (DNNs) has become the foundation technique for many safety and security critical artificial intelligence (AI) applications such as autonomous driving, medical
imaging and biometric authentication~\cite{yuan2014droid,shen2017deep,kim2017interpretable}. No doubt to say, reliability and safety consideration is the biggest concern for these sensitive applications, and it is imperative to mitigate any possible threats.

One of the main threats all DNNs face is adversarial examples (AEs), which are
carefully crafted adversarial inputs to deceive the model to make a wrong decision. Due to the severeness of this problem, there has been a large body of research on AE attacks~\cite{szegedy2013intriguing,goodfellow2014explaining,papernot2016limitations,carlini2017towards,moosavi2016deepfool} and defenses~\cite{kurakin2016adversarial,tramer2017ensemble,papernot2016distillation,ross2018improving} from both academia and industry.

However, as pointed out in~\cite{lingdeepsec}, existing defense techniques can only defend against limited types of attacks under restricted settings. It is very difficult to have a universal solution to defend against all possible adversarial example attacks, especially considering the fact that we are only able to explore a small set of data domains in the deep learning training phase while the remaining large unexplored space could be exploited by attackers.
In practice, the impacts of different misclassifications caused by adversarial example attacks may vary significantly. Considering a medical image diagnostic system, missing a life threatening disease is usually regarded by a patient as much more severe than a false positive diagnosis. Thus, it is imperative to take the impacts of different adversarial example attacks into consideration and develop a configurable defense mechanisms to satisfy the unique requirements of different applications, which has not be explored in the literature.

In this paper, our idea is to refine the loss function of the DNNs by adding a new term, called \emph{attack sensitive matrix}, to perceive the costs of different targeted attacks. Then, by adjusting the attack sensitive matrix in the loss function, we are able to manipulate the defense strength for different targeted attacks, thereby effectively increasing the attack effort for those high-cost attacks. 
To the best of our knowledge, this is the first configurable defense mechanism against targeted AE attacks. The main contributions of this paper include:
\begin{itemize}
\item We propose a novel configurable defense method for adversarial example attacks by refining the loss function of DNNs during training.
\item We present two common defense objectives that can be achieved by our configurable defense: one is to increase the weighted average robustness; while the other is to increase the lower bound robustness of the system.
\item We conduct the experiments on CIFAR-10 and GTSRB data sets and show that our solution can achieve significant improvement compared to the state-of-the-art defense methods under a range of attacks.
\end{itemize}

The remainder of this paper is organized as follows: First, we introduce some preliminary knowledge, the related work and motivation in Section~\ref{sec:preliminary}. Next, we detail the proposed configurable defense mechanism by refining the loss function in Section~\ref{sec:method}. After that, in Section~\ref{sec:objective}, two efficient algorithms are presented to achieve two common defense objectives. Lastly, we show the experimental results in Section~\ref{sec:results} and conclusions in Section~\ref{sec:conclusions}.

\section{Preliminaries and Motivation}\label{sec:preliminary}

\subsection{Adversarial Example Attack}

There are two kinds of adversarial example attacks: \emph{targeted attacks}~\cite{szegedy2013intriguing,papernot2016limitations,carlini2017towards,chen2017ead} and \emph{un-targeted attacks}~\cite{moosavi2016deepfool,moosavi2017universal,he2018decision}. Targeted attacks try to make DNNs misclassify the input from a correct source label to a targeted malicious label, while the objective of un-targeted attacks is to fool DNNs make mistakes, regardless of the targeted label.
In this paper, we focus on the configurable defense against the targeted adversarial example attack, which can be formulated as follows:
\begin{equation}
A_{i,j}: \mathop{\arg\min}_{\Delta \boldsymbol x_{ij}}\  \ F(\boldsymbol x_i + \Delta \boldsymbol x_{ij})=j,\  \ i\neq j.
\end{equation}
Under the attack $A_{i,j}$, the DNN model $F$ misclassifies the sample $\boldsymbol x_{ij}$ from the source correct label $i$ to the targeted malicious label $j$. The objective of attackers is to find the minimum perturbation vector $\Delta \boldsymbol x_{ij}$ added on the input so that $\boldsymbol x_i$ is misclassified as $j$. 
For a classifier with $n$ classes, there are $n*(n-1)$ kinds of targeted adversarial example attacks, and each may cause different costs for system users.





The robustness under the adversarial example attack is defined as the correct classification rate of adversarial examples, which measures the authenticity of the model under attacks. The higher the correct classification rate, the more robust the DNN model. The mathematical formulation of the model robustness under the attack $A_{i,j}$ is as follows:

\begin{equation}
R_{i,j} = \frac {\# (F(\boldsymbol x_i+\Delta \boldsymbol x_{ij})=i) } {\#(\boldsymbol x_i+\Delta \boldsymbol x_{ij})},
\end{equation}
where the numerator denotes the number of adversarial examples generated under the targeted attack $A_{i,j}$ that can still be correctly classified as $i$. The denominator represents the total number of adversarial examples crafted under the attack $A_{i,j}$. 


\subsection{Related Work}

In the literature, there are two categories of defenses against adversarial example attacks: one is to build detection systems to recognize adversarial examples during model usage~\cite{ma2018characterizing,xu2017feature,meng2017magnet}, and the other tries to train more robust models to successfully classify adversarial examples~\cite{szegedy2013intriguing,papernot2016distillation}. For detecting adversarial examples, a certain number of techniques have been explored, such as performing statistical tests~\cite{ma2018characterizing} or training an additional model for detection~\cite{meng2017magnet}. However, as adversarial examples are very close to legitimate samples, it has been shown that many detection methods can be bypassed by attackers easily~\cite{carlini2017adversarial}. For training a more robust model,
defensive distillation~\cite{papernot2016distillation} first builds a classification model and smoothes its softmax layer by dividing a constant, then trains a robust model with the soft labels that are the outputs from the first one. Input gradient regularization~\cite{ross2018improving,gu2014towards} method targets to train a contractive network, minimizing the model input gradients towards the output predictions. Adversarial training~\cite{szegedy2013intriguing, madry2017towards} augments the original training set with crafted adversarial examples.
However, all the previous defenses attempt to provide a universal solution, improving the model robustness against all kinds of adversarial example attacks, which is impossible~\cite{lingdeepsec}.

\begin{figure}[t]
\includegraphics[width=1\columnwidth]{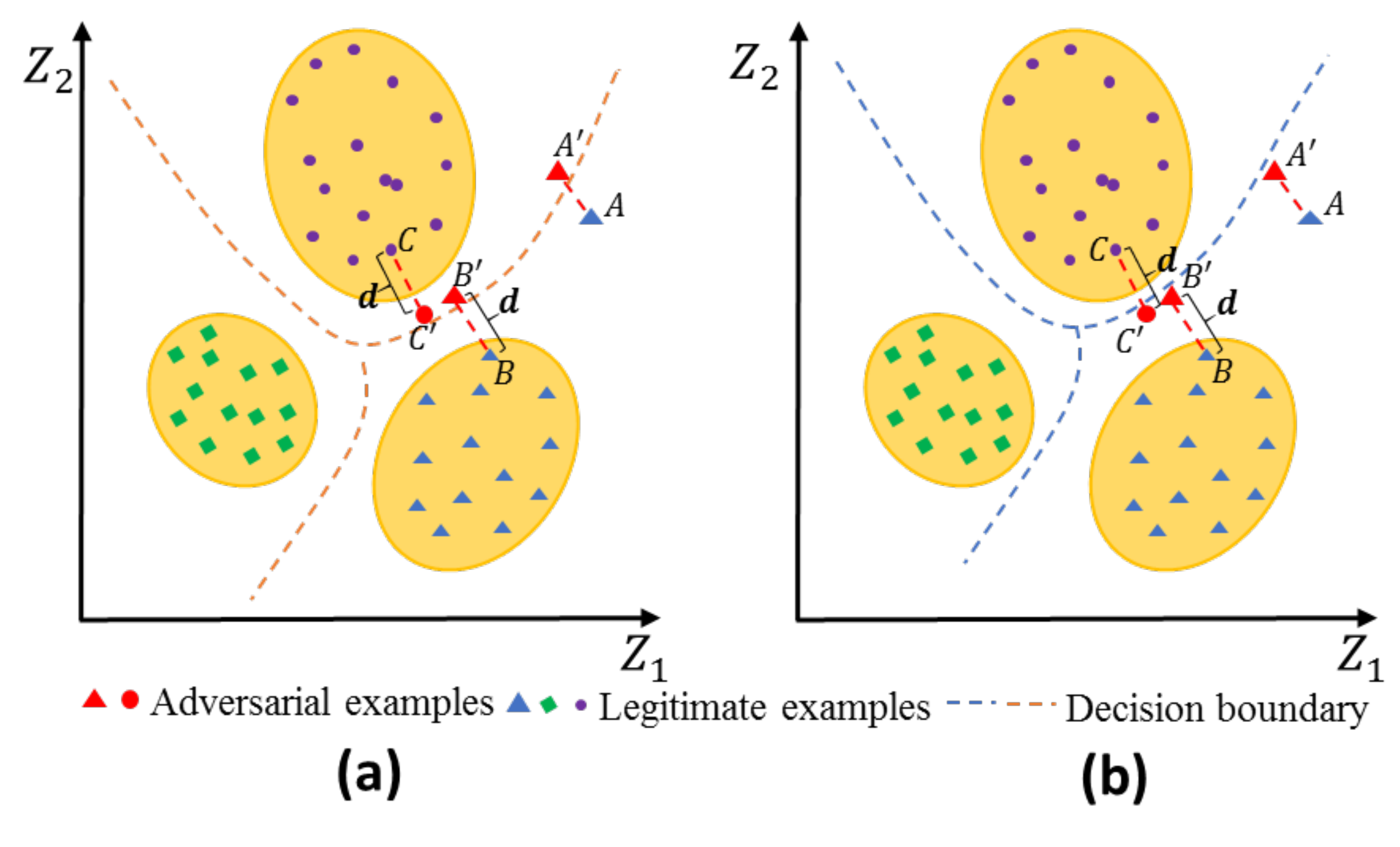}
\caption{Moving decision boundary to mitigate adversarial examples $A'$ and $B'$ would lead to the misclassification of $C'$. }\label{fig:motivation}
\vspace{-15pt}
\end{figure}

\subsection{Motivation}
As the minimum distance from legitimate inputs of a class to the decision boundary dictates the amount of perturbations required to generate the corresponding adversarial examples. Moving the decision boundary of a classifier to mitigate some threats would inevitably leads to some other new threats. For example, in Figure~\ref{fig:motivation}, the classifier in Figure~\ref{fig:motivation}(a) correctly classifies the adversarial example $C'$, while misclassifies adversarial examples $A'$ and $B'$. However, when moving the decision boundary as shown in Figure~\ref{fig:motivation}(b), the classifier will correctly classify adversarial examples $A'$ and $B'$, but misclassify $C'$. Therefore, it is very difficult, if not impossible, to have a universal solution to defend against all possible adversarial example attacks.

In fact, the impacts of different adversarial example attacks vary significantly in a particular machine learning system. For example, considering a traffic sign classifier used in self-driving cars, it will not cause a problem to misclassify a ``yield'' sign as a ``stop'' sign, but it may cause severe traffic accidents for the opposite misclassification. Therefore, a preferred defense mechanism should protect the DNN model in such a manner that it is more difficult for attackers to perturb a legitimate 'stop' sign to be misclassified as other road signs when compared to other possible misclassifications. In other words, it is imperative to have a configurable defense mechanism against adversarial example attacks to satisfy the unique requirement in particular machine learning applications.

Motivated by the above, in this paper, we propose to investigate the configurable defense mechanisms for satisfying the unique requirements of different applications, as a universal solution is usually intractable.
To the best of our knowledge, this is the first work on configurable defense against adversarial example attack, as detailed in the following sections.

\section{The Proposed Method}\label{sec:method}
The idea of our configurable defense is to devise a configurable loss function that includes a matrix to perceive the costs of different misclassifications caused by different adversarial example attacks. In this section, we first introduce the limitation of the cross entropy loss used in training DNNs. Then, we propose our configurable loss functions to mitigate its limitation. Lastly, we present the overflow of our configurable defense, performing adversarial training with the refined loss functions to achieve a robust model.

\subsection{Limitation of Cross Entropy Loss}

In the classification problems based on DNNs, the most popular loss used is the cross entropy loss~\cite{Goodfellow-et-al-2016}, which has shown great success in achieving high classification accuracy. Assume there are $n$ classes, the cross entropy loss for one training sample is:
\begin{equation}
L_{cross}= - \sum_{i=1}^n y_i*log(\hat{y}_i),
\end{equation}
where $y_i$ is the $i$-th element in the one-hot encoded format of the true label. If the sample is of $i$ class, $y_i$ equals to 1, otherwise, it is zero. $\hat{y_i}$ is the probability of class $i$ predicted by the classifier. As a result, the cross entropy loss for a sample is simplified as $-y_t*log(\hat{y_t})$, in which $t$ is the index of the true label for this sample.
Based on this fact, we observe that the cross entropy loss only cares about the prediction accuracy for the true class, regardless of others. Thus, it cannot consider the costs of misclassifications caused by different adversarial example attacks. 

\subsection{Attack Sensitive Loss}
To solve this problem, we propose a refined loss, called \emph{attack sensitive loss}, which incorporates all the prediction probability of classes instead of only considering the prediction probability of the true class. Then, we introduce an \emph{attack sensitive matrix} to the loss, which perceives the costs of different targeted attacks for configuring the attack strength. In this paper, we propose two formulations for the attack sensitive loss.

The first one is defined as follows:
\begin{equation}
L_{sensitive}^1= \sum_{i=1}^n (1-y_i) * \hat{y}_i * M_{t,i},
\end{equation}
where $(1-y_i) * \hat{y}_i$ calculates the error magnitude of class $i$. As training samples are labeled with one-hot format, when a sample is not class $i$, then $1-y_i$ equals to 1. But the classifier erroneously predicts this sample to class $i$ with probability $\hat{y}_i$. The higher the $\hat{y}_i$, the larger the prediction error. $M_{t,i}$ is a value in the attack sensitive matrix $M$. It denotes the costs of targeted attack $A_{t,i}$. The larger $M_{t,i}$, the larger the loss caused by $A_{t,i}$ during training. Then the model will become more robust against $A_{t,i}$ attack after well trained. In this way, our defense can configure the model robustness against different adversarial example attacks by adjusting the attack sensitive matrix $M$.

%
%

The second attack sensitive loss is slightly different from the first one in the way of calculating the error magnitude. It has the following format:
\begin{equation}
L_{sensitive}^2= \sum_{i=1}^n (\hat{y}_i- \hat{y}_{t}) * M_{t,i}\\.
\end{equation}
The error magnitude of class $i$ in this formulation is calculated as the gap between the predicted probability of class $i$ and the true class $t$, denoted as $\hat{y}_i- \hat{y}_{t}$. Apparently, the larger the probability gap, the larger the error made by this model. Then, we multiply the error magnitude with the corresponding attack sensitive value $M_{t,i}$. Similarly, we can configure the attack sensitive matrix $M$ to adjust the costs caused by different misclassifications. 


\subsubsection{Loss Function Combination}
As the cross entropy loss achieves good accuracy in training DNNs and our attack sensitive loss can configure the model robustness against different adversarial example attacks, we combine these two loss functions together to achieve a good tradeoff between the model classification accuracy and robustness.
The combined loss function is formulated as follows:
\begin{equation}
L_{configurable}= L_{cross} + \lambda * L_{sensitive},
\end{equation}
where $L_{cross}$ is the cross entropy loss, and $L_{sensitive}$ is the attack sensitive loss. $\lambda$ is a parameter to balance the effects of these two losses. 


\subsection{Overflow of Our Configurable Defense}

\begin{figure}[h]
\begin{flushright}
\includegraphics[width=0.9\columnwidth]{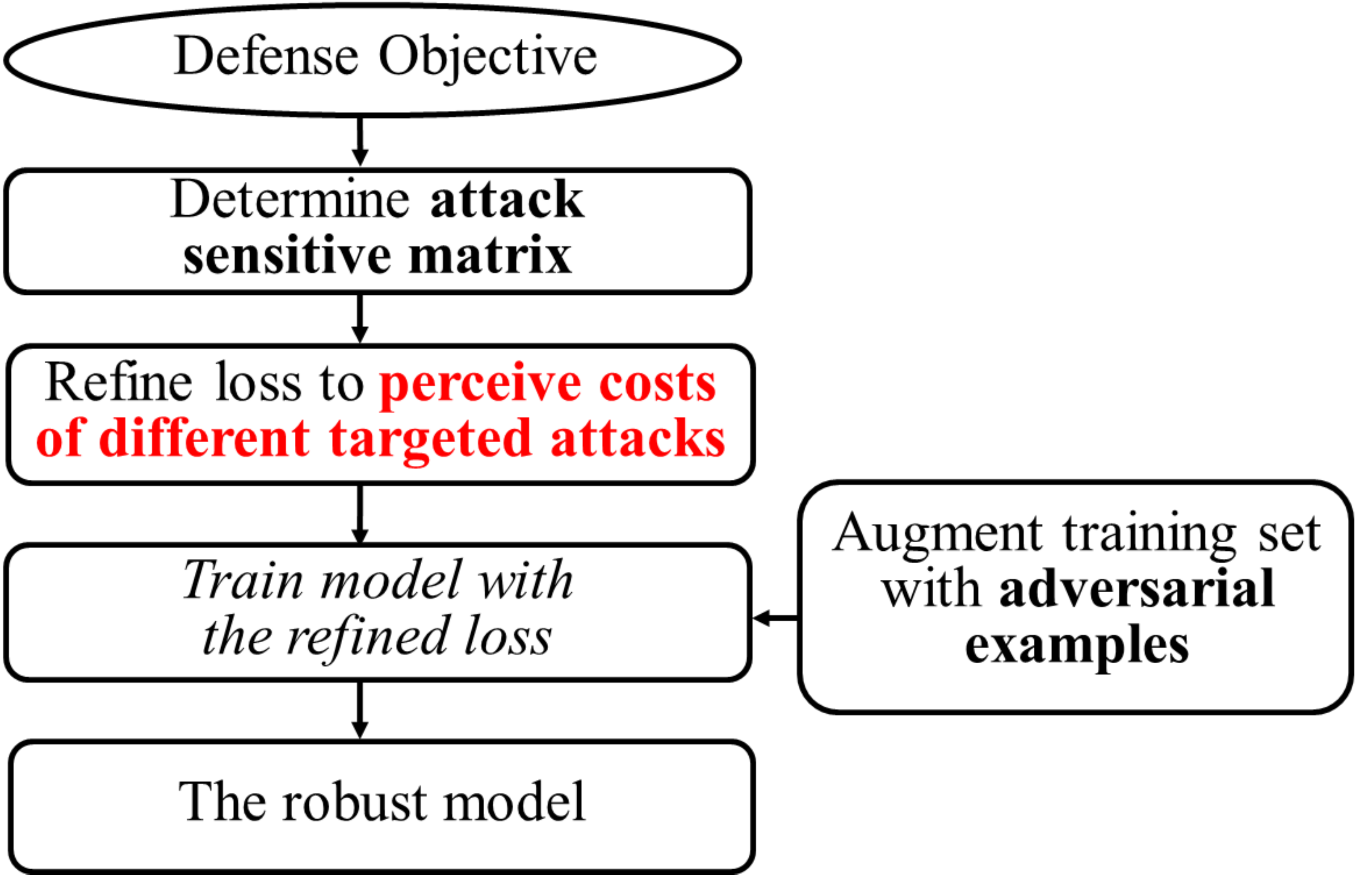}
\caption{The procedure of our configurable defense. }\label{fig:overview}
\end{flushright}
\end{figure}

The idea of the proposed configurable defense is to introduce our loss functions into adversarial training method, achieving different defense strength for satisfying the unique requirements of different applications. The overflow of our configurable defense is shown in Figure~\ref{fig:overview}, where we first determine the attack sensitive matrix based on the system objective. Then we refine the loss functions of DNNs by including the attack sensitive matrix as an important parameter to perceive costs of different targeted attacks.
 Finally, we train with the refined loss on the augmented training set and obtain the robust model satisfying the application requirement.



The key challenge in our configurable defense is to assign the attack sensitive matrix $M$ properly. As too large values in $M$ will degrade the performance for legitimate samples, while too small values in $M$ will not bring a satisfactory robustness improvement.
To solve this challenge, we propose algorithms to optimize the attack sensitive matrix, achieving an appropriate robustness improvement under different defense objectives. The details are demonstrated in the following section.

\section{Defense Objectives}\label{sec:objective}
In this section, we introduce two common defense objectives that can be achieved by the proposed configurable mechanism: one is to increase the weighted average robustness for systems, which concentrates more on some particular severe targeted attacks; while the other is to increase the lower bound robustness of the system considering the bottleneck security.
\subsection{Increase Weighted Average Robustness}
Many machine learning systems have robustness preference, such as in disease diagnosis systems, the sick cases should not be attacked to become healthy ones. As a result, the defense techniques should concentrate more on the robustness against the high-cost targeted attacks, instead of treating all attacks equally. Based on this analysis, we define the system robustness as the weighted average robustness: 
\begin{equation}
	\overline R=\sum_{i=1}^n \sum_{j=1, j \ne i}^n R_{i,j}*W_{i,j},
\end{equation}
where $R_{i,j}$ is the robustness under the targeted attack $A_{i,j}$. $W_{i,j}$ is the weight or the cost of the successful attack $A_{i,j}$ specified by system users. The more important $R_{i,j}$ in the system robustness, the larger $W_{i,j}$.

As discussed earlier, increasing the values in the attack sensitive matrix will increase the model robustness against the corresponding targeted attacks, however it will inevitably influence the classification rate of legitimate samples. In order to ensure the system usability, the accuracy of legitimate samples should be constrained. Then we can increase the system robustness whenever possible under this constraint. Overall, we formulate the problem as follows: 
\begin{equation}
     \begin{gathered}
        \mathop{\arg\max}_{M}\  \overline R\\
        s.t. \quad Accu(legitimate) > \xi.\\
     \end{gathered}
\end{equation}
Our target is to find the appropriate attack sensitive matrix $M$ in our loss functions that can maximize the system robustness when the accuracy of legitimate samples is greater than a given threshold $\xi$.

\begin{algorithm}[h]\label{alg:train}
{\fontsize{10pt}{10pt}\selectfont
\KwIn{Training set $\rho$, validation set $v$, minimum required accuracy $\xi$, weight $W_{A_{(i,j)}}$, step $\Delta$.
}

\KwOut{Attack sensitive matrix $M$.}
Initialize the attack sensitive matrix $M$ with all 1s except 0s in diagonal;\\

    $T$ $\leftarrow$Sort $(i,j)$ according to $W_{i,j}$ in descending order;\\
    \For{$\forall t_k \in T$} {
    $terminate = false$;\\
        \While{ $terminate == false$}{

        Train the model on $\rho$ with the refined loss;\\
        $Accu$ $\leftarrow$ Evaluate the accuracy for $v$;\\
            \If{$Accu > \xi$}{
            $M_{t_k} = M_{t_k} + \Delta$; \\
            }
            \Else{
            $terminate = true$;\\
            $M_{t_k} = M_{t_k} - \Delta$;
            }
        }
    }
    Return $M$.
\caption{Increase Weighted Average Robustness.}\label{alg:weighted}
}
\end{algorithm}

However, finding the optimal attack sensitive matrix for this problem is not easy, as the values in this matrix are not constrained and the search space is infinite. Besides, there are no signs, such as gradients for minimizing loss functions, to guide us to update the matrix. To solve this problem, we propose a simple yet efficient greedy algorithm to find an appropriate attack sensitive matrix. The intuition is that increasing the value $M_{i,j}$ will certainly increase the robustness against the targeted attack $A_{i,j}$ when the model is well trained with our refined loss. As a result, we first increase the value of $M_{i,j}$ that corresponds to the most serious attack, which is the attack with the largest $W_{i,j}$ defined by users. When $M_{i,j}$ is too large to violate the constraint, we fix $M_{i,j}$ and start to increase the value in $M$ corresponding to the second serious attack. This process is continued until we cannot increase any value in $M$ under the accuracy constraint.

The detailed process of finding an appropriate attack sensitive matrix is shown in Algorithm~\ref{alg:weighted}, in which we first initialize $M$. Then, in line 2, we sort the tuple $(i,j)$ in descending order according to the attack seriousness $W_{i,j}$ given by users and store the tuples in $T$. Next, the algorithm traverses all positions in the attack sensitive matrix in descending order of the attack seriousness. For each value in $M$, we train the model with our new losses and evaluate the accuracy of legitimate samples in line 6-7. If the constraint on the accuracy of legitimate samples is satisfied, the element of $M$ in position $t_k$ will be increased for a small value $\Delta$ in line 8-10. The process is continued until the constraint is violated and then the flag of termination is set and the last update of $M_{t_k}$ is restored. 


\subsection{Increase Lower Bound Robustness}
Apart from increasing the weighted average robustness, there are some cases where users care more about the lower bound robustness. As the robustness under different targeted attacks are extremely imbalanced in traditionally trained models, some attacks are really easy to implement, while some are quite difficult to succeed~\cite{papernot2016limitations}. The lower bound robustness is the bottleneck of the system security, as a result, it is essential to improve the lower bound robustness. In this paper, we formulate the problem as follows:

\begin{equation}
     \begin{gathered}
        \mathop{\arg\max}_{M}\  \min (R)\\
        s.t. \quad Accu(legitimate)> \xi.\\
     \end{gathered}
\end{equation}
Our objective is to find the appropriate attack sensitive matrix $M$ that can maximize the lower bound robustness when the accuracy of legitimate samples is greater than a given threshold.

Similarly, we propose an efficient algorithm to solve this problem. In previous sections, we know that increasing $M_{i,j}$ can improve the robustness under the target attack $A_{i,j}$. Therefore, we can increase the element $M_l$ in the attack sensitive matrix corresponding to the lower bound robustness until the constraint is violated. As the position of $M_l$ in $M$ may change frequently between iterations, the convergence speed may be degraded due to these kinds of oscillations. Thus, we propose to increase a batch of elements in $M$ corresponding to the lowest robustness simultaneously in each iteration. 


\begin{algorithm}[h]\label{alg:train}
{\fontsize{10pt}{10pt}\selectfont
\KwIn{Training set $\rho$, validation set $v$, minimum required accuracy $\xi$, batch size $t$, step $\Delta$.}

\KwOut{Attack sensitive matrix $M$.}
Initialize the attack sensitive matrix $M$ with all 1s except 0s in diagonal;\\
$terminate = false$;\\
\While{$terminate == false$}{
        Train the model on $\rho$ with the refined loss;\\
        $Accu$ $\leftarrow$ Evaluate the accuracy for $v$;\\
        \If{$Accu > \xi$}{
            Add $\Delta$ to $t$ elements in $M$ corresponding to the lowest robustness; \\
        }
        \Else{
            $terminate = true$;\\
            Minus $\Delta$ from $t$ elements in $M$ corresponding to the lowest robustness;
        }


}
Return $M$.
\caption{Increase lower bound robustness.}\label{alg:lowest}
}
\end{algorithm}

The whole process is listed in Algorithm~\ref{alg:lowest}, where we first initialize the attack sensitive matrix the same as in Algorithm~\ref{alg:weighted}. Then in each iteration, we first train the model with our refined loss functions and obtain the accuracy of samples in the validation set in line 4-5. If the new accuracy does not violate the constraint, we choose the $t$ elements in $M$ corresponding to the $t$ lowest robustness and increase them by a small value $\Delta$. However, if the accuracy violates the constraint, we set the terminate flag to become true and decrease $\Delta$ from the $t$ elements in $M$ that have been updated in the last iteration in line 10-11.

\begin{figure*}[!htb]
\centering
\subfigure[\textbf{C\&W}]{
\includegraphics[scale=0.19]{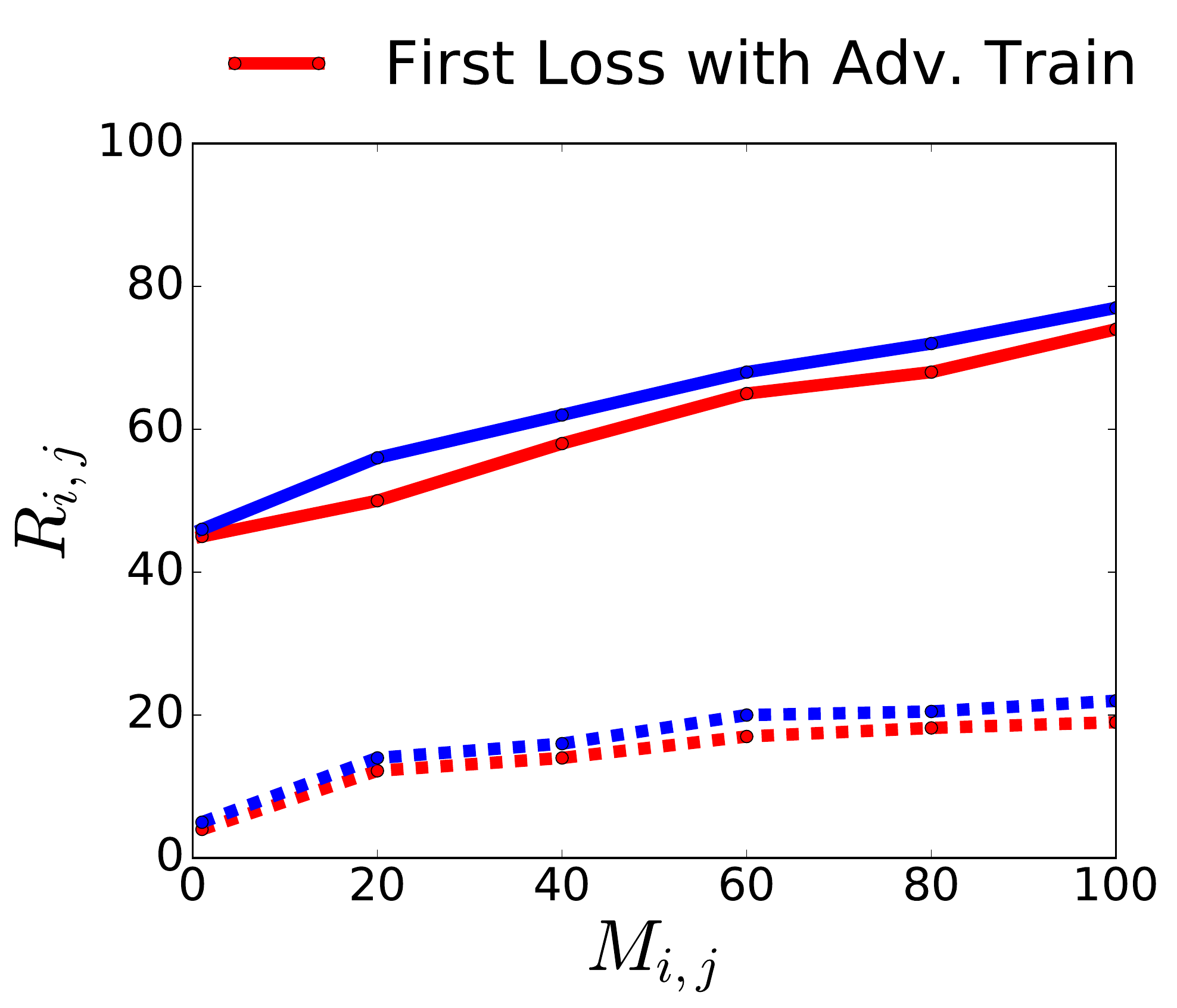}}
\subfigure[\textbf{IFGSM}]{
\includegraphics[scale=0.19]{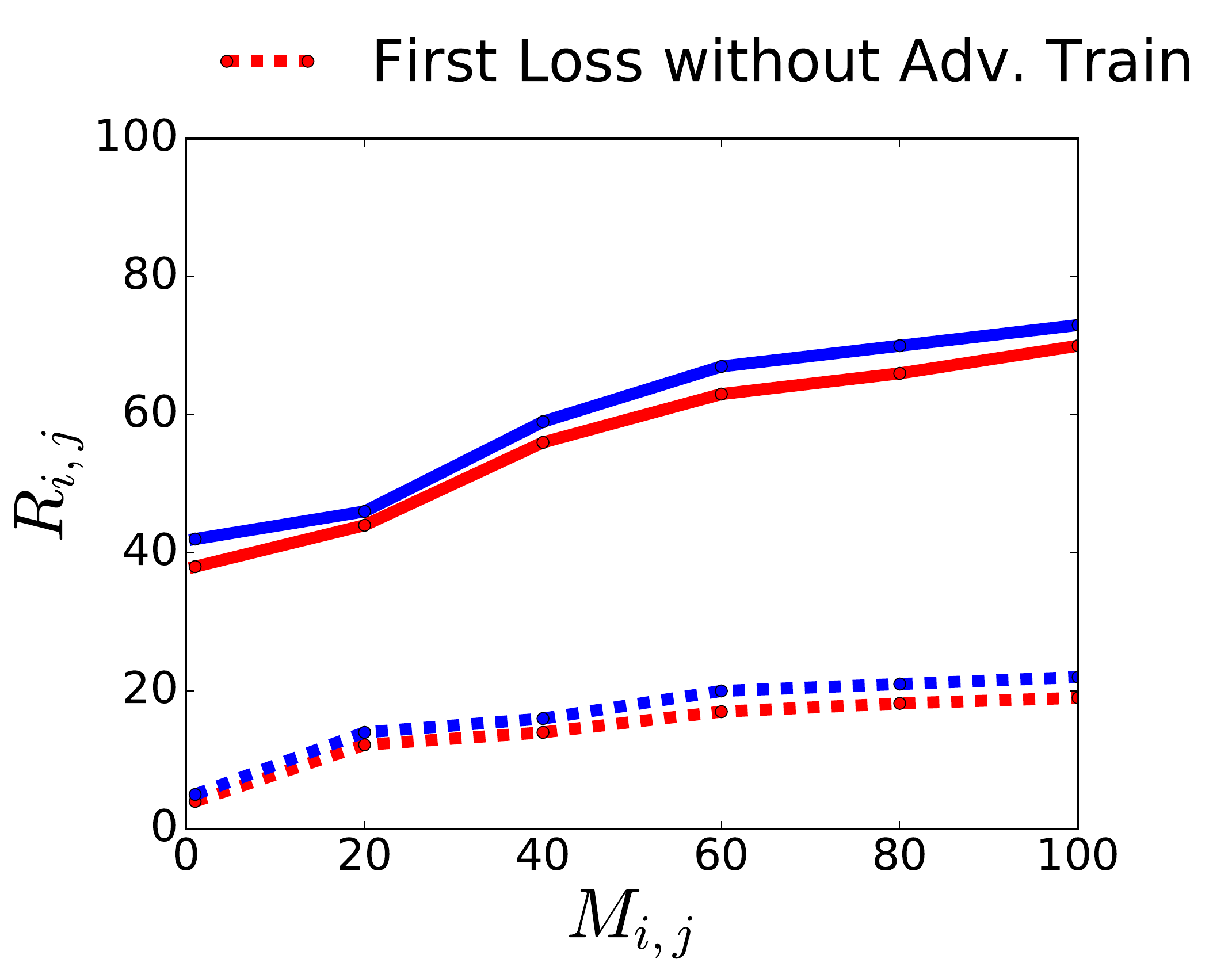}}
\subfigure[\textbf{PGD}]{
\includegraphics[scale=0.19]{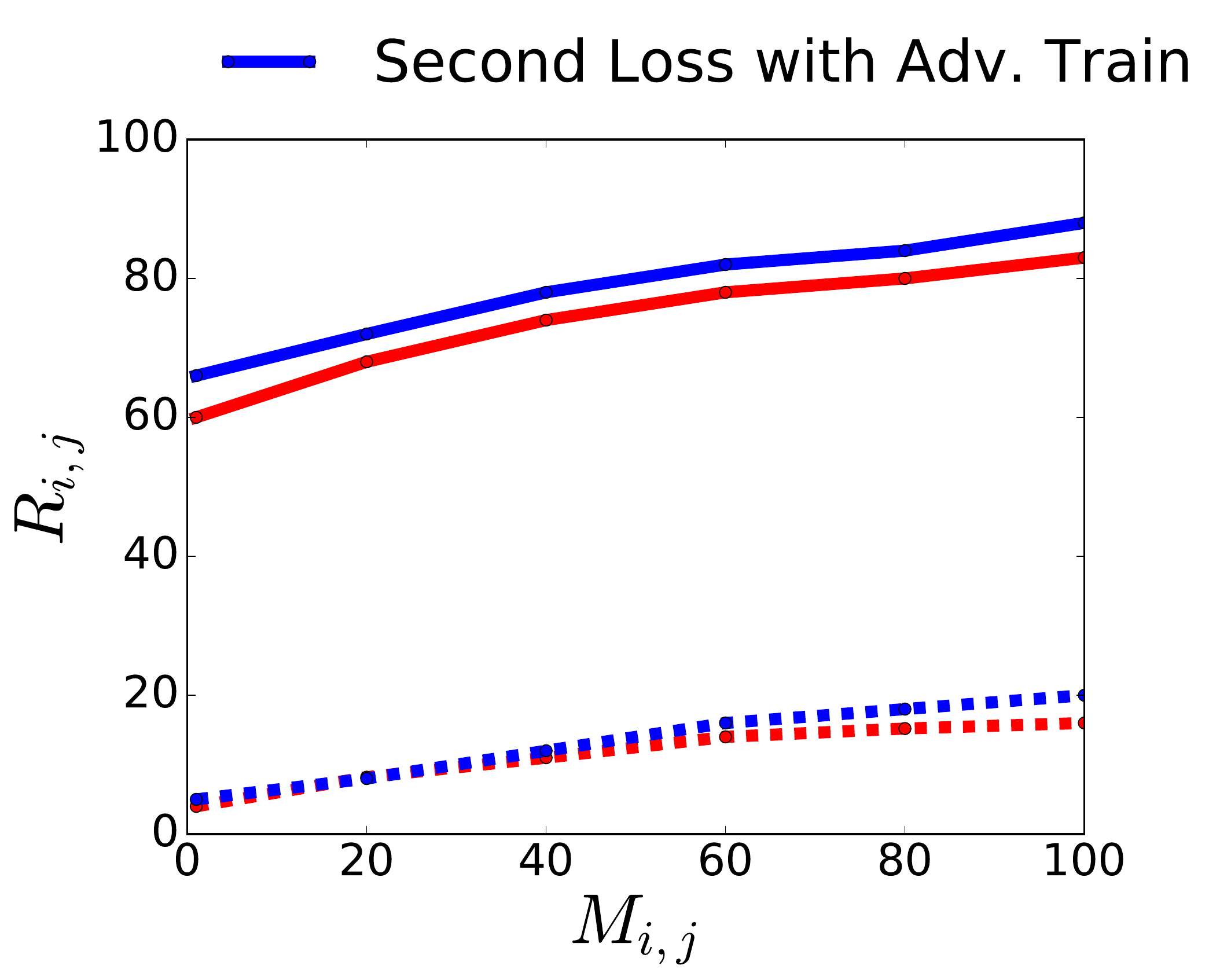}}
\subfigure[\textbf{Legitimate Samples Accuracy}]{
\includegraphics[scale=0.19]{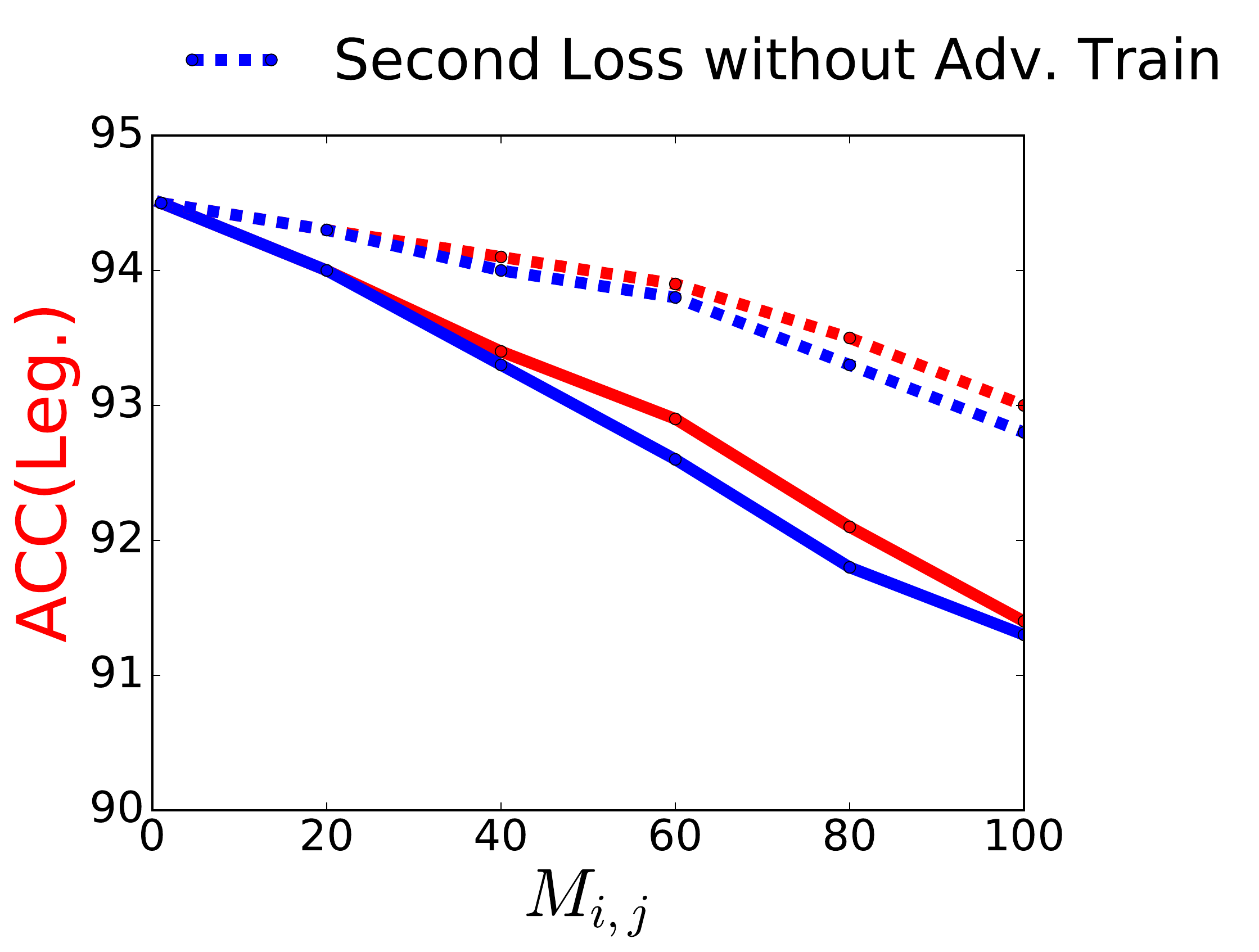}}
\caption{The impacts of the attack sensitive matrix on the model accuracy for adversarial examples and legitimate samples, where we set $\lambda=1$ in our losses.} 
\label{fig:matrix}
\end{figure*}

\section{Experimental Results}\label{sec:results}
In this section, we evaluate the effectiveness of our proposed configurable defense from three aspects: the first one is to evaluate the configurability of our refined loss functions on the model robustness; the second is the performance of the two proposed algorithms for different defense objectives; the last one is to evaluate the performance of our configurable defense on a practical problem (road sign recognition system).

\subsection{Experimental Setup}

\textbf{Datasets:} Our experiments are performed on CIFAR-10~\cite{krizhevsky2014cifar} and GTSRB (German Traffic Sign Recognition Benchmark)~\cite{stallkamp2012man} data sets.
The CIFAR-10 contains 60000 color images that represent 10 different natural objects. Each image has the size of 32*32*3. The GTSRB data set contains 50000 images representing 43 kinds of road signs.
The intensity values of pixels in all these images are scaled to a real number in the range of $[0,1]$.


\textbf{DNN Models:} The model architectures for these data sets are deep convolutional neural networks (CNNs). They achieve 94.8\% and 98.8\% classification accuracy for CIFAR-10 and GTSRB, respectively, which is comparable to the state-of-the-art results.

\textbf{Baselines:} In the experiments, we compare the performance of our configurable defense mechanism with three state-of-the-art defenses:
\begin{itemize}
\item  \emph{PGD-based Adversarial Training}~\cite{madry2017towards}, it trains with the training set augmented with adversarial examples generated by PGD attack;


\item \emph{Feature Squeezing}~\cite{xu2017feature}, it improves the model robustness by reducing the bit depth of inputs;

\item \emph{Input Gradient Regularization}~\cite{ross2018improving}, it optimizes the model for more smooth input gradients based on the predictions during training.
\end{itemize}
We use three state-of-the-art attacks, PGD~\cite{madry2017towards}, IFGSM~\cite{kurakin2016adversarial} and C\&W~\cite{carlini2017towards}, to evaluate the performance of these defenses.

\subsection{Configurability of Loss Functions}

In this section, we evaluate the configurability of our refined loss functions on the model robustness for a specific targeted attack by adjusting the corresponding attack sensitive value in $M$. 
In the experiments, we randomly choose a value $M_{i,j}$ to change, then we train models with our two configurable losses by increasing $M_{i,j}$ from 1 to 100. The model robustness $R_{i,j}$ is evaluated under IFGSM, C\&W and PGD attacks, respectively. We also evaluate the prediction accuracy of legitimate samples without attacks when training with our new losses. The results for CIFAR-10 data set are shown in Figure~\ref{fig:matrix}. To better evaluate performance of our configurable loss functions, we train the models with two schemes: one in training without adversarial examples, and the other is adversarial training (augmenting the training set with adversarial examples crafted by PGD).



%

\begin{table*}[!htb]
\centering
\begin{center}
\scalebox{0.82}[0.82]{
\setlength{\arrayrulewidth}{0.8pt}
\begin{tabular}{|c|c|c|c|c|c|}
 \hline
 &\multirow{2}{*}{PGD-based Adv. Train}&\multirow{2}{*}{Input Gradient Regularization}&\multirow{2}{*}{Feature Squeezing}&\multicolumn{2}{c|}{Our Configurable Defense}\\
 \cline{5-6}
& & & &Train with PGD Adv.& Train with Ensemble Adv.\\
 \hline

IFGSM &  63.4\%&67.2\%&63.7\%& 81.1\%&90.8\%\\
\hline
PGD&  68.6\%&63.6\%&58.9\%& 86.5\% &88.1\%\\
\hline
C\&W& 55.4\%&60.1\%&52.1\%& 76.4\% &84.3\%\\
\hline
\end{tabular}
}
\caption{The performance of our configurable defense compared with the state-of-the-art defenses for improving the weighted average robustness $\overline R$ on CIFAR-10.}\label{tb:average}
\end{center}
\end{table*}

\begin{table*}[!htb]
\centering
\begin{center}
\scalebox{0.82}[0.82]{
\setlength{\arrayrulewidth}{0.8pt}
\begin{tabular}{|c|c|c|c|c|c|}
 \hline
&\multirow{2}{*}{PGD-based Adv. Train}&\multirow{2}{*}{Input Gradient Regularization}&\multirow{2}{*}{Feature Squeezing}&\multicolumn{2}{c|}{Our Configurable Defense}\\
 \cline{5-6}
 & & & &Train with PGD Adv.& Train with Ensemble Adv.\\
 \hline
 IFGSM&  46.2\%&53.5\%&43.1\%& 63.7\% &71.4\%\\
\hline
PGD & 44.6\%&50.3\%&42.8\%& 65.1\% &66.2\%\\
\hline
C\&W& 39.4\%&44.7\%&36.9\%& 56.3\% &64.8\%\\
\hline
\end{tabular}
}
\caption{The performance of our configurable defense compared with the state-of-the-art defenses for improving the lower-bound robustness $min(R)$ on CIFAR-10.}\label{tb:lower}
\end{center}
\end{table*}

Firstly, from the dotted lines in Figure~\ref{fig:matrix}(a), (b), (c), we can see that our refined loss functions can improve the model robustness even without adversarial training. The model robustness $R_{i,j}$ increases from about 5\% to 18\% when $M_{i,j}$ increases from 1 to 100. However, the accuracy of legitimate samples degrades about 1.5\%, as the dotted lines in Figure~\ref{fig:matrix}(d) shows.

Secondly, from the solid lines in Figure~\ref{fig:matrix}(a), (b), (c), with adversarial training, our losses are effective to increase the model robustness $R_{i,j}$ against the targeted attack by increasing the corresponding value $M_{i,j}$. But when $M_{i,j}$ increases largely, the accuracy of legitimate samples degrades about 3.2\%. This indicates that increasing model robustness would inevitably degrade the accuracy of legitimate samples. We need to consider this when achieving our defense objectives.


Finally, we can observe that our second loss function performs better than the first one, as its improvement of model robustness is larger. We analyze this phenomenon that our second loss uses the probability gap to denote the error magnitude. In this way, the model trained with this loss tends to learn the unique features in each class to increase the probability gap, and thus becomes more robust.

\subsection{Defense Objective Evaluation}
In this section, we evaluate the effectiveness of the algorithms proposed for the two different defense objectives, improving the weighted average robustness and the lower bound robustness. As our second refined loss performs better than the first one, these results are conducted by the second loss function. To compare the performance of different defense mechanisms, we control that the degradation of legitimate sample accuracy is less than 1\%. That is, we set $\xi$ to 93.8\% for CIFAR-10.


\subsubsection{Increase Weighted Average Robustness}
We evaluate the effectiveness of our Algorithm~\ref{alg:weighted} for improving the weighted average robustness, compared with the three baseline defense mechanisms. The weights of $R_{i,j}$ in $\overline R$ for each targeted attack are set as follows: we randomly select 6 weights and set them to 0.4, 0.2, 0.08, 0.06, 0.04 and 0.02, respectively. Except for $W_{i,i} = 0$, the remaining weights are set as the same value, which is determined under the constraint that all weights are sum to 1. In practice, the weights are set based on application requirements.


Table~\ref{tb:average} shows the weighted average robustness achieved by different defense methods under a range of attacks on CIFAR-10.
We can see that the Input Gradient Regularization achieves the best weighted average robustness among the three baseline defenses under IFGSM and C\&W attacks, with robustness of 67.2\% and 60.1\%, respectively. For the PGD attack, PGD-based Adversarial Train achieve the best performance of 68.6\%. This is commensurate with the previous discovery that adversarial training is not effective for the attacks, based on which adversarial examples generated are not included in the training set.

Our configurable defense can even improve this benefit. When training with adversarial examples generated by PGD, we get 81.1\%, 86.5\% and 76.4\%  weighted average robustness under IFGSM, PGD and C\&W attacks, respectively.
And when we use the ensemble adversarial training method (including adversarial examples crafted by IFGSM, PGD and C\&W), our configurable defense can achieve 90.8\% and 84.3\% weighted average robustness under IFGSM and C\&W attacks, which is almost 35\% improvement compared with the best results among the three baselines.
The reason of this significant improvement of our defense is that we judiciously protect the model from severe attacks instead of treating all of them equally as previous methods do. 


\begin{figure*}[!htb]
\centering
\subfigure[Original Image]{
\includegraphics[scale=0.85]{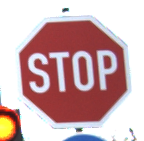}}
\subfigure[PGD Adversarial Train]{
\includegraphics[scale=0.85]{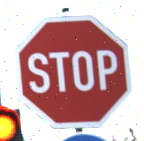}}
\subfigure[Gradient Regularization]{
\includegraphics[scale=0.85]{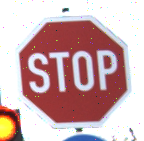}}
\subfigure[Feature Squeezing]{
\includegraphics[scale=0.85]{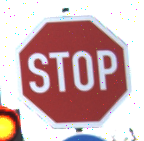}}
\subfigure[Configurable Defense]{
\includegraphics[scale=0.85]{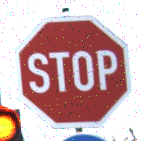}}
\caption{ Original sample and the corresponding adversarial examples crafted against defended models by different defenses under C\&W attack.} 
\label{fig:road}
\end{figure*}

\begin{table*}[!htb]
\centering
\begin{center}
\scalebox{0.82}[0.82]{
\setlength{\arrayrulewidth}{0.8pt}
\begin{tabular}{|c|c|c|c|c|c|}
 \hline
&\multirow{2}{*}{PGD-based Adv. Train}&\multirow{2}{*}{Input Gradient Regularization}&\multirow{2}{*}{Feature Squeezing}&\multicolumn{2}{c|}{Our Configurable Defense}\\
 \cline{5-6}
& & & &Train with PGD Adv.& Train with Ensemble Adv.\\
 \hline
IFGSM &58.3\%&61.2\%& 56.1\% &80.7\%&87.8\%\\
\hline
PGD& 63.8\%&58.1\%& 53.4\% &84.8\%&86.2\%\\
\hline
C\&W&53.1\%&56.3\%& 51.6\% &76.8\%&82.6\%\\
\hline
\end{tabular}
}
\caption{our configurable defense compared with the state-of-the art defenses under a range of attacks on GTSRB.}\label{tb:road}
\vspace{-15pt}
\end{center}
\end{table*}

\subsubsection{Increase Lower Bound Robustness}
We evaluate the effectiveness of our Algorithm~\ref{alg:lowest} for improving the lower bound robustness, compared with the three baseline defense mechanisms.
Table~\ref{tb:lower} shows the lower bound robustness achieved by different defense methods on CIFAR-10. 
The best performance among the three baseline defenses is achieved by the Input Gradient Regularization method. The lower bound robustness is 53.5\%, 50.3\% and 44.7\% under IFGSM, PGD and C\&W attacks, respectively. However, compared with our configurable defense, this improvement is not good enough. Our solution with the PGD adversarial training and ensemble training can achieve 22\% and 30\% improvement compared to the best results obtained by the three baselines.

To conclude, our solution makes significant improvement compared with previous defense methods for different defense objectives with only little degradation on the accuracy of legitimate samples. When a universal defense solution is not available, it is essential to employ our configurable defense mechanism to protect the model against those severe attacks.

\subsection{Case Study}\label{use-case}
To evaluate the effectiveness of our configurable defense on practical problems, we implement a use case on traffic road sign recognition system. We conduct the experiments on the GTSRB data set. In a road sign recognition system, the most important security guarantee is that the ``Stop'' sign should not be classified as others. Thus, the defense objective in this system is to improve the model robustness against the adversarial example attack from misclassifying the ``Stop'' sign into other labels ($A_{Stop, Non-stop}$).

This problem can be solved using Algorithm~\ref{alg:weighted}, improving the weighted average robustness of the model, where the weight of $R_{Stop,Non-stop}$ in $\overline R$ should be the largest. In this experiment, we set $W_{Stop,Non-stop}$=0.8 and the left weights are set as the same value except for $W_{i,i}$. All the weights are sum to 1. Therefore, in this setting, the adversarial example attacks that cause misclassifying the ``Stop'' sign to other ``Non-stop'' will incur the most serious impact.


Table~\ref{tb:road} shows the weighted average robustness of different defense methods under the attack $A_{Stop, Non-stop}$. The maximum degradation of legitimate sample accuracy is 1\%. 
We observe that the best robustness under IFGSM attack among three baselines is achieved by Input Gradient Regularization, which is 61.2\%. However, our configurable defense can even improve this benefit. We can correctly classify 80.7\% and 87.8\% adversarial examples on the stop signs under IFGSM when using PGD based adversarial training and ensemble training, respectively. 
The results for PGD and C\&W attacks are similar, that our configurable defense can largely improve the robustness compared with the best results achieved by the three baselines.

Figure~\ref{fig:road} shows the adversarial examples crafted against different defended models under the C\&W attack. The first image is the original sample, and the following three are the adversarial examples crafted against models defended with the three baseline methods. The final figure is the adversarial example generated against the model defended by our configurable defense. We can see that the perturbations needed by our method are the largest compared with the three baselines. This corresponds to the conclusions in Table~\ref{tb:road}, that our configurable defense can improve system robustness considering application requirements by largely increasing the attack strength.

\section{Conclusions}\label{sec:conclusions}
In this paper, we propose a configurable defense against adversarial example attacks by refining loss functions during training, adding a new term to perceive the cost of different target attacks. In this way, the model robustness can be configured by adjusting the attack sensitive matrix in our new losses. Moreover, we present two efficient algorithms to achieve two different defense objectives: one is to increase the weighted average robustness, and the other is to increase the lower bound robustness. Experimental results on CIFAR-10 and GTSRB data sets show that the proposed mechanism can significantly achieve different defense objectives when compared with the state-of-the-art techniques.

{
\bibliographystyle{ieee}
\bibliography{ref}
}

\end{document}